\pdfoutput=1

\documentclass[11pt]{article}

\usepackage[final]{acl}

\usepackage{cmap}           
\usepackage{times}
\usepackage{latexsym}
\usepackage[T1]{fontenc}
\usepackage[utf8]{inputenc}
\input{glyphtounicode}
\usepackage{microtype}
\usepackage{inconsolata}
\usepackage{graphicx}

\usepackage{colortbl}

\usepackage{multirow}
\usepackage{array}
\usepackage{tabularx}

\newcolumntype{Y}{>{\centering\arraybackslash}X}
\newcolumntype{L}{>{\RaggedRight\arraybackslash}X}
\usepackage{xspace}     
\usepackage{booktabs}   
\usepackage{float}      
\usepackage{amsmath}
\usepackage{amssymb}
\usepackage[capitalise,nameinlink]{cleveref}
\usepackage{xcolor}
\usepackage{ragged2e}    
\usepackage[breakable,skins]{tcolorbox}  
\usepackage{soul}                   
\usepackage[noend]{algpseudocode}   
\usepackage{algorithm}              
\usepackage{placeins}               

\definecolor{PBorder}{HTML}{E69F00}
\definecolor{PBg}{HTML}{FBE4C7}
\definecolor{QBorder}{HTML}{56B4E9}
\definecolor{QBg}{HTML}{CBE7F7}

\definecolor{AAWMP}{HTML}{E69F00}
\definecolor{AAWMPbg}{HTML}{FBE4C7}
\definecolor{AAWMQ}{HTML}{56B4E9}
\definecolor{AAWMQbg}{HTML}{CBE7F7}

\newtcolorbox{casebox}[1]{%
  enhanced, breakable,
  width=\columnwidth,
  colback=white, colframe=black!60,
  boxrule=0.5pt, arc=1mm,
  title=#1, colbacktitle=black!5, coltitle=black!90,
  fonttitle=\small\bfseries,
  segmentation style={dashed, draw=black!30},
  left=4pt, right=4pt, top=4pt, bottom=4pt,
  fontupper=\small\RaggedRight}

\newtcolorbox{aawmcase}[1][]{%
  breakable,
  width=\columnwidth,
  colback=white,
  colframe=black!35,
  boxrule=0.4pt,
  arc=1pt,
  left=4pt,right=4pt,top=4pt,bottom=4pt,
  boxsep=0.5pt,
  fontupper=\RaggedRight,
  #1
}

\newtcolorbox{promptbox}[1]{%
  enhanced,
  breakable, 
  title={\small #1},
  colback=white, colframe=black!40, colbacktitle=black!8,
  coltitle=black, fonttitle=\small\bfseries,
  boxrule=0.4pt, arc=1pt,
  left=4pt, right=4pt, top=4pt, bottom=4pt,
  fontupper=\footnotesize\RaggedRight}

\providecommand{\stdv}[1]{\raisebox{-0.2ex}{{\scalebox{0.8}{\tiny$\pm$}}{\tiny#1}}}

\newif\ifanonymousawardcopy
\anonymousawardcopyfalse

\title{Beyond Next-Observation Prediction:\\
Agent-Authored World Modeling for Sequential Decision Making}

\ifanonymousawardcopy
\author{\textbf{Anonymous EMNLP 2026 Submission}}
\else
\author{
\textbf{Guangfeng Cai\textsuperscript{1}},
\textbf{Kaibing Yang\textsuperscript{1}},
\textbf{Shuo He\textsuperscript{2}},
\textbf{Yu Li\textsuperscript{1}},\\
\textbf{Shengtian Yang\textsuperscript{1}},
\textbf{Jiaqi Lv\textsuperscript{1}},
\textbf{Lei Feng\textsuperscript{1}\thanks{Corresponding author.}}\\
\textsuperscript{1}Southeast University, Nanjing, China\\
\textsuperscript{2}Meituan, Shanghai, China\\
\texttt{\{cgfeng,fenglei\}@seu.edu.cn}
}
\fi

\begin{document}
\maketitle

\begin{abstract}

Recent studies on world modeling for Large Language Model (LLM) agents typically formulate the learning objective as next-observation prediction.
However, this objective ties supervision to what a transition happens to reveal, which may omit the dynamics most relevant to the agent's current decision.
To bridge this gap, we propose Agent-Authored World Modeling (AAWM), a training procedure that constructs supervision from the policy's own decision needs.
Specifically, at each state, the agent identifies what it needs to understand about the environment before acting.
These needs drive the retrieval of relevant transition evidence across trajectories, which is then synthesized into training targets that capture decision-oriented dynamics instead of reconstructing the next observation.
This aligns the training objective with the dynamics the policy needs before acting, not with the contents of the next observation.
Experimental results validate the effectiveness of AAWM across multiple environments and training settings.
These results show that decision-aware world-model targets provide a more effective learning signal than next-observation prediction.

\end{abstract}

\section{Introduction}
\label{sec:intro}

LLM agents~\citep{openai2024gpt4technicalreport, kimiteam2026kimik2openagentic, glm2024chatglmfamilylargelanguage, deepseekai2025deepseekv3technicalreport} have increasingly operated in partially observed environments that require multi-turn interactions to gather information and complete tasks.
In such settings, success depends not only on choosing fluent actions but also on representing environment dynamics: which actions are valid, which state variables persist, and which observations matter for the next decision.
Recent work therefore fine-tunes large language models to predict the environment's next response, treating this objective as world modeling~\citep{zhang2025alee, li2025fromwordtoworld}.
The underlying assumption is that \textit{better prediction of the next environment output will produce better action selection.}

\begin{figure}[t]
\centering
\includegraphics[width=\columnwidth]{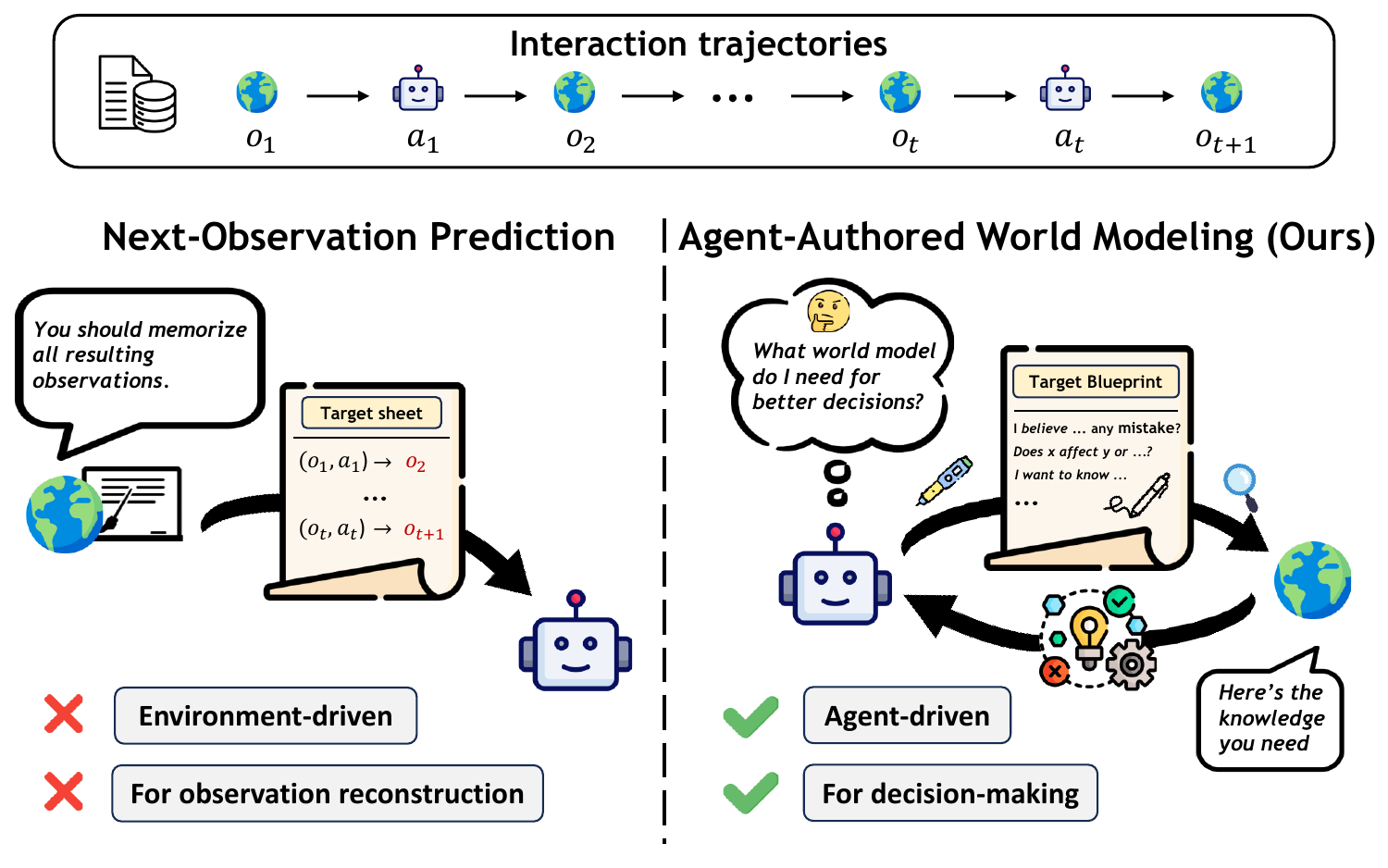}
\caption{Comparison between next observation prediction and \textsc{AAWM}.
Next-observation prediction trains on the observation returned by each action, so the target is determined by what the environment happens to reveal.
\textsc{AAWM} asks the policy what it needs to know before acting and writes targets that describe the dynamics needed for decision making.}
\label{fig:intro}
\end{figure}

This assumption overlooks a key difference between prediction and decision:

the policy often needs the state information that affects its next action rather than a full reconstruction of the next observation~\citep{nair2020gap}.
Moreover, \citet{predictingacting2024} observed an empirical trade-off between next-token prediction and goal-directed agent behavior, suggesting that prediction quality and decision quality are not the same objective.
This raises a core target selection problem: \textbf{\textit{what environment dynamics does the policy need in the world modeling target before acting?}}

Prior work has mostly addressed this problem by changing the form of the target.
Some methods compress observations into semantic future representations~\citep{semanticwm2025}, summarize transition-level state changes~\citep{wma2025}, or add task information for downstream planning~\citep{wkm2024}.
These targets are often more compact or more planning-friendly than raw next observations, but their content is still determined by the transition exposed in the dataset.

As a result, the training target remains environment-driven rather than decision-driven: \textbf{\textit{it reflects what the environment happened to reveal, instead of what the policy needs to understand before acting.}}

To address this problem, we propose \textsc{AAWM} (Agent-Authored World Modeling), a training procedure that constructs training targets from the policy's current decision needs.
\Cref{fig:intro} illustrates the difference from next observation prediction.
At each state, the policy first articulates its beliefs about environment dynamics and open questions whose answers could change the next action.
These statements serve as retrieval queries over a pool of transition records, gathering evidence about the queried dynamics from other trajectories.
The retrieved evidence, together with the current transition and the policy's statements, is then synthesized into a natural-language target describing decision-oriented dynamics.
Fine-tuning on these targets encourages the policy to represent the dynamics that matter for action selection instead of only reconstructing the next observation.

We evaluate AAWM on ALFWorld and WebShop at two model scales.
With the same imitation learning and reinforcement learning setup, AAWM consistently outperforms next-observation world modeling, with gains of up to 6.3 and 6.2 success-rate points on the two environments.
A separate AgentGym evaluation~\citep{xi2025agentgym} across four environments confirms that AAWM is the only world modeling initialization that improves over imitation learning in every setting.
Component ablations show that Self-Probing, Transition Retrieval, and world modeling fine-tuning each contribute to the final gain.
Training analysis further shows that AAWM-initialized policies sustain broader decision-oriented reasoning during reinforcement learning while task success continues to rise.

Our contributions are summarized as follows:
\begin{itemize}
\vspace{-2.5mm}
\item We identify target selection as a central problem in world modeling for language agents: training targets should capture the dynamics that affect action choice, not only reconstruct environment responses.
\vspace{-2.5mm}
\item We instantiate this principle as \textsc{AAWM}, a world modeling procedure in which the policy's own beliefs and uncertainties determine what the training target addresses.
\vspace{-2.5mm}
\item We demonstrate that AAWM outperforms next-observation world modeling across two environments, two model scales, and both supervised and reinforcement learning settings.
\end{itemize}

\begin{figure*}[t]
\centering
\includegraphics[width=\textwidth]{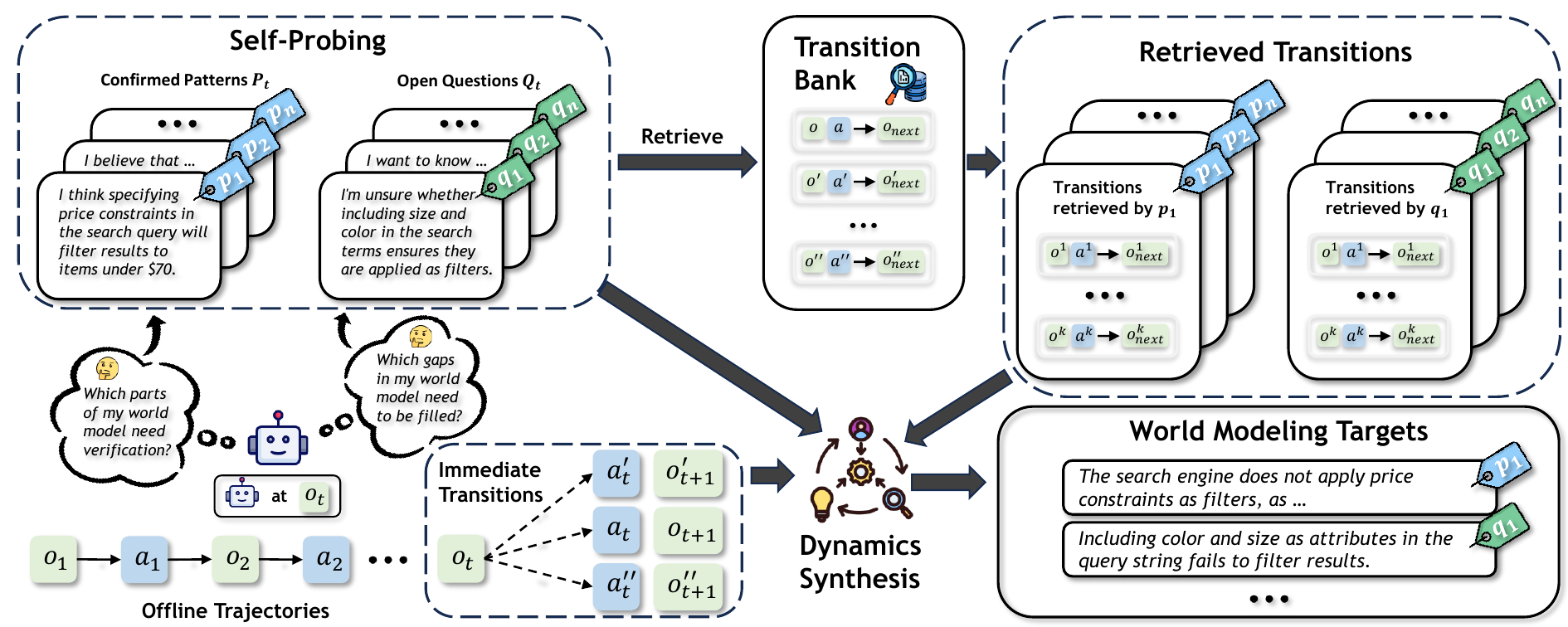}
\caption{\textsc{AAWM} target construction pipeline.
At a decision context $o_t$, the policy produces confirmed patterns $P_t$ and open questions $Q_t$ through Self-Probing.
Each proposition retrieves evidence from the transition bank $\mathcal{T}$, and the immediate transitions $\mathcal{I}_t$ provide local evidence from the same context.
Dynamics Synthesis combines these inputs into world modeling targets that correct mistaken beliefs and answer open questions when evidence permits.}
\label{fig:pipeline}
\end{figure*}

\section{Related Work}
\label{sec:related-work}

\paragraph{LLMs as decision-making agents.} 
LLM agents have been widely studied for sequential decision making in software engineering~\citep{yang2024sweagentagentcomputerinterfacesenable}, embodied interaction~\citep{wang2023voyageropenendedembodiedagent}, web navigation~\citep{gur2024realworldwebagentplanninglong}, and GUI operation~\citep{zhang2024lookscreensmultimodalchainofaction,hong2024cogagentvisuallanguagemodel}. 
Prompting methods such as chain-of-thought reasoning and ReAct~\citep{yao2023reactsynergizingreasoningacting} structure intermediate reasoning traces to decompose complex tasks, while self-reflection mechanisms such as Reflexion~\citep{shinn2023reflexionlanguageagentsverbal} incorporate verbal feedback from prior failures to revise action selection. 
Memory~\citep{park2023generativeagentsinteractivesimulacra}, retrieval~\citep{nakano2022webgptbrowserassistedquestionansweringhuman}, and tool-use~\citep{gou2024toratoolintegratedreasoningagent,schick2023toolformerlanguagemodelsteach} mechanisms further extend agent capabilities by supplying external context or executable operations during interaction.
Recent training methods also reuse successful tool transitions, route different behavioral phases to specialized experts, and extract learning signals from rollout groups in which every attempt fails~\citep{li2026phgpo,yang2026phase,yang2026progress}.
These approaches improve policy learning for long-horizon tasks.
AAWM complements this line of work by constructing world modeling targets from the policy's decision needs.

\paragraph{World models for LLM agents.}
World models have been widely used to provide future information and auxiliary supervision for sequential decision making.
Recent work across language and vision-language decision-making systems has instantiated this idea through transition-level state change summarization~\citep{wma2025}, semantic compression~\citep{semanticwm2025}, result state prediction~\citep{guo2025dymo}, action simulation~\citep{dynathink2025}, and knowledge injection~\citep{wkm2024}.

A related line of work involves the agent more directly in world-model construction.
\citet{spa2025} internalizes environment dynamics from the agent's own rollouts;
\citet{wang2025vagenreinforcingworldmodel} reinforces explicit prediction of environment states during multi-turn interactions;
\citet{agent2world2025} reconstructs the thinking process with world-model simulation; and
\citet{zhang2025alee} generate additional interaction data from the agent's own actions and train on the resulting outcomes.

Despite this progress, the content of the training target is still determined by what each transition happens to reveal, which may omit the dynamics most relevant to the agent's current decision. 
Our work therefore aims to dynamically adjust the policy's target to its current beliefs and open questions.

\section{Preliminaries}
\label{sec:method}

\textbf{Problem setup.}
We consider a partially observed text environment
$\mathcal{E}$ in which a language-model policy $\pi_\theta$ interacts
through textual observations and actions.
At step $t$, the policy receives the decision context $o_t$ and samples
an action $a_t\sim\pi_\theta(\cdot\mid o_t)$.
The environment returns the next observation $o_{t+1}$, producing
trajectories $\mathcal{D}=\{\tau_i\}$ with
$\tau_i=(o_0,a_0,o_1,\ldots,o_T)$, and we write
$\mathcal{U}(\mathcal{D})=\{(o_t,a_t,o_{t+1})\}$ for the set of all
one-step transition records.
From these records we maintain two collections.
The \emph{global transition pool} $\mathcal{T}$ contains records from all trajectories as a single retrievable collection.
The \emph{immediate transition set} $\mathcal{I}_t\subseteq\mathcal{T}$ contains records originating from the same context $o_t$, including the
logged transition and additional transitions obtained by executing alternative actions from that context.

\paragraph{World modeling objective.}
A standard text world model predicts the environment response conditioned on the current context and action.
For each transition record, a textual outcome target
$\bar{o}_{t+1}=g(o_{t+1})$ is constructed from the next observation, where $g$ may be the identity map, a state change extractor, or a summarizer.
The next response prediction objective is to minimize
\begin{equation}
\mathcal{L}_{\mathrm{obs}}(\theta)
=
-\sum_{\substack{(o_t,a_t,o_{t+1})\\\in\,\mathcal{U}(\mathcal{D})}}
\log p_\theta\!\left(\bar{o}_{t+1}\mid o_t,a_t\right)
  \label{eq:obs-wm}
\end{equation}
This objective is useful when the next observation contains the dynamics
needed for control, but the dynamics that determine a good action can be
sparse, implicit, or distributed across earlier transitions.
Before acting, the policy may need to verify whether its current
understanding of the environment is correct and to resolve specific
gaps in that understanding.
These needs reflect the state of the policy's own world model, not the
content of the next observation.
AAWM addresses this by constructing world modeling targets from the
policy's decision-oriented modeling needs.
These targets provide an auxiliary training objective for the same
model that is subsequently trained and evaluated on downstream agentic tasks.

\section{Agent-Authored World Modeling}
\label{sec:method:overview}
To align the world modeling objective with what the policy needs before acting, 
\textsc{AAWM} constructs targets from the policy's current decision context,
with an overview of the pipeline shown in \Cref{fig:pipeline}.

\textit{Self-Probing} elicits the policy's current beliefs and open questions.
\textit{Transition Retrieval} selects supporting records from $\mathcal{T}$.
\textit{Dynamics Synthesis} combines the probes, the immediate transitions, and the retrieved transitions into a fine-tuning target.
The following subsections describe these stages in detail.

\subsection{Self-Probing}
\label{sec:method:selfprobing}

Instead of defining target content from the next observation,
\textsc{AAWM} elicits the policy's modeling needs at context $o_t$.
Specifically, the policy is prompted to produce two sets of propositions:
\begin{equation}
  (P_t,Q_t)\sim\pi_\theta(\cdot\mid o_t),
  \label{eq:selfprobe}
\end{equation}
where $P_t=\{p_t^1,\ldots,p_t^n\}$ are \emph{confirmed patterns} and
$Q_t=\{q_t^1,\ldots,q_t^n\}$ are \emph{open questions}.

Confirmed patterns specify parts of the policy's current understanding that may require verification or correction against retrieved transitions.

Open questions identify dynamics that the policy has not yet resolved and whose answers could affect the next action.

Together, $(P_t,Q_t)$ determine what the training target should contain and drive the subsequent retrieval and synthesis steps.
The prompt template is in Appendix~\ref{app:prompts}.

\subsection{Transition Retrieval}
\label{sec:method:targets}

Since the required dynamics may be absent from the immediate transition set $\mathcal{I}_t$,
we retrieve relevant transitions from the global
transition pool $\mathcal{T}$.

For each proposition $r\in P_t\cup Q_t$, AAWM retrieves $K$ transitions from $\mathcal{T}$ using maximal marginal relevance \citep{carbonell1998mmr}, which balances relevance to $r$ with diversity across the selected records, ensuring the evidence covers different aspects of the queried dynamics instead of redundant records.
The retrieved set is
\begin{equation}
  \mathcal{R}_t =
  \bigcup_{r\in P_t\cup Q_t}
  \mathrm{Retrieve}(r,\mathcal{T};K).
  \label{eq:retrieve}
\end{equation}
Implementation details are in Appendix~\ref{app:retrieval}.

\subsection{Dynamics Synthesis}
\label{sec:method:synthesis}

Given the policy's acting needs specified by the probes and the transition evidence grounded by retrieval, the final stage synthesizes an agent-authored world-modeling target.

Specifically, an external instruction-following model
$f_{\mathrm{syn}}$ receives the current context $o_t$, confirmed patterns $P_t$, open questions $Q_t$, the immediate transitions $\mathcal{I}_t$ from the same context, and the retrieved transitions $\mathcal{R}_t$ from the global pool:
\begin{equation}
  y_t =
  f_{\mathrm{syn}}\!\left(
  o_t,\ P_t,\ Q_t,\ \mathcal{I}_t,\ \mathcal{R}_t
  \right).
  \label{eq:synth}
\end{equation}
The model is instructed to correct any unsupported beliefs, resolve open questions the evidence permits, and summarize the dynamics most relevant to action
selection.
The resulting $y_t$ thus describes environment dynamics shaped by what the policy needs before acting.
The synthesis prompt is in Appendix~\ref{app:prompts}, and the synthesis model is described in \Cref{sec:exp:setup}.

\subsection{World Modeling}
\label{sec:method:wmft}

Let $\mathcal{D}_{\mathrm{A}}=\{(o_t,y_t)\}$ denote the dataset produced by Dynamics Synthesis.
AAWM fine-tunes the policy with the agent-authored world-modeling objective

\begin{equation}
  \mathcal{L}_{\mathrm{A}}(\theta)
  =
  -\sum_{(o_t,y_t)\in\mathcal{D}_{\mathrm{A}}}
  \log p_\theta\!\left(y_t\mid o_t\right),
  \label{eq:aawm-wm}
\end{equation}
where the subscript $\mathrm{A}$ denotes AAWM.
The target therefore updates the policy toward the dynamics specified by
its own decision needs.
The resulting parameters initialize the subsequent imitation learning
and reinforcement learning stages.

\section{Experiments}
\label{sec:exp}

\subsection{Setup}
\label{sec:exp:setup}

\paragraph{Environments and evaluation.}
We evaluate on ALFWorld~\citep{shridhar2020alfworld}
and WebShop~\citep{yao2022webshop}.
ALFWorld is a textual embodied environment with six household task types, Pick, Clean, Cool, Look, Heat, and Pick2.
WebShop is an HTML shopping environment where the agent searches, inspects, and purchases products according to user instructions.
ALFWorld reports success rate for each task type and the aggregate \textbf{All}, while WebShop reports success rate \textbf{Succ.} and reward score \textbf{Score}.
We follow the environment configuration and evaluation protocol from the open-source repository~\citep{feng2025gigpo}.

\paragraph{Baseline.}
We use Implicit World Modeling (\textsc{IWM}) as the world modeling baseline, a representative next observation prediction method for language agents~\citep{zhang2025alee}.
Given the current observation and action $(o_t,a_t)$, \textsc{IWM} trains the model to predict the next observation.
Following the original IWM setting, the target is the raw textual observation on ALFWorld and an offline summary of the next observation on WebShop.

\paragraph{World modeling data.}
We draw world modeling data from AgentTraj-L~\citep{xi2025agentgym} and keep all successful trajectories executable in the open-source repository~\citep{feng2025gigpo} as $\mathcal{D}$.
At every action step, we execute the logged action and three additional actions from the same state, producing four transitions that form $\mathcal{I}_t$ and populate $\mathcal{T}$.
Both \textsc{IWM} and \textsc{AAWM} share the same $\mathcal{D}$ and $\mathcal{T}$ within each backbone scale.
We construct both sets only from training trajectories, so evaluation trajectories never enter target construction.
Retrieval also excludes records from the query episode.
Trajectory counts and action-sampling details are provided in Appendix~\ref{app:main-exp-details}.

\paragraph{Models and target construction.}
We train the policy at two scales, Qwen2.5-1.5B-Instruct and Qwen2.5-7B-Instruct~\citep{qwen2025qwen25technicalreport}.
Self-Probing uses the policy backbone at the corresponding scale; in the main experiments, it produces $n=3$ confirmed patterns and $n=3$ open questions per state. Transition Retrieval embeds records with Qwen3-Embedding-0.6B~\citep{qwen3embedding}.
We use Qwen3-30B-A3B-Instruct-2507~\citep{yang2025qwen3} only for target construction, where it produces WebShop observation summaries for \textsc{IWM} and synthesized targets for \textsc{AAWM}.
This keeps the target construction model shared between the two world modeling methods.

\paragraph{Training protocol.}
Both \textsc{IWM} and \textsc{AAWM} first initialize the policy by supervised fine-tuning on their respective world modeling targets, followed by a lightweight imitation learning stage on 32 trajectories from $\mathcal{D}$ for one epoch.
All RL rows then run Group Relative Policy Optimization (GRPO; \citealp{shao2024deepseekmath}).

\Cref{tab:main} reports prompting baselines and trained-policy results at each backbone scale.
Within each backbone, all trained rows share the same trajectory source, transition pool, imitation data, and GRPO budget.

Base denotes downstream training without world modeling initialization.
We report mean and standard deviation over three seeds.
More details are in Appendix~\ref{app:main-exp-details}.

\subsection{Main Results}
\label{sec:exp:main}
\begin{table*}[!t]
\centering\footnotesize

\newcommand{\aawmPanelRow}[1]{%
  \multicolumn{11}{@{}c@{}}{\textbf{#1}} \\
}
\newcommand{\aawmBackboneRule}{\specialrule{0.25pt}{1pt}{1pt}}

\resizebox{.99\textwidth}{!}{%
\begingroup
\setlength{\tabcolsep}{2.8pt}
\begin{tabular}{@{}l l ccccccc c c@{}}
\toprule
\multirow{2}{*}{Backbone} & \multirow{2}{*}{Method}
 & \multicolumn{7}{c}{ALFWorld} & \multicolumn{2}{c}{WebShop} \\
\cmidrule(lr){3-9}\cmidrule(lr){10-11}
 & & Pick & Look & Clean & Heat & Cool & Pick2 & All & Score & Succ. \\
\midrule
\aawmPanelRow{Prompting}
\midrule
\multirow{2}{*}{Closed-source Models}
  & GPT-4o
  & 75.3 & 60.8 & 31.2 & 56.7 & 21.6 & 49.8 & 48.0
  & 31.8 & 23.7 \\
  & Gemini-2.5-Pro
  & 92.8 & 63.3 & 62.1 & 69.0 & 26.6 & 58.7 & 60.3
  & 42.5 & 35.9 \\
\aawmBackboneRule
\multirow{3}{*}{Qwen2.5-1.5B-Instruct}
  & Qwen2.5
  &  5.9 &  5.5 &  3.3 &  9.7 &  4.2 &  0.0 &  4.1
  & 23.1 &  5.2 \\
  & ReAct
  & 17.4 & 20.5 & 15.7 &  6.2 &  7.7 &  2.0 & 12.8
  & 40.1 & 11.3 \\
  & Reflexion
  & 35.3 & 22.2 & 21.7 & 13.6 & 19.4 &  3.7 & 21.8
  & 55.8 & 21.9 \\
\aawmBackboneRule
\multirow{3}{*}{Qwen2.5-7B-Instruct}
  & Qwen2.5
  & 33.4 & 21.6 & 19.3 &  6.9 &  2.8 &  3.2 & 14.8
  & 26.4 &  7.8 \\
  & ReAct
  & 48.5 & 35.4 & 34.3 & 13.2 & 18.2 & 17.6 & 31.2
  & 46.2 & 19.5 \\
  & Reflexion
  & 62.0 & 41.6 & 44.9 & 30.9 & 36.3 & 23.8 & 42.7
  & 58.1 & 28.8 \\
\midrule
\aawmPanelRow{Imitation Learning initialized by World Modeling}
\midrule
\multirow{3}{*}{Qwen2.5-1.5B-Instruct}
  & Base
  & 18.4\stdv{3.6} &  6.8\stdv{2.4} & 13.7\stdv{5.8}
  & 11.6\stdv{3.8} &  8.8\stdv{4.8} &  3.7\stdv{2.1}
  & 12.0\stdv{2.6}
  & \textbf{26.6}\stdv{4.9} &  4.2\stdv{1.2} \\
  & \textsc{IWM}
  &  5.9\stdv{3.4} &  1.6\stdv{1.1} &  3.2\stdv{1.8}
  &  2.4\stdv{1.5} &  1.5\stdv{1.0} &  0.7\stdv{0.5}
  &  3.1\stdv{1.2}
  & 21.6\stdv{3.3} &  2.1\stdv{0.5} \\
\rowcolor{gray!15}
  & \textbf{\textsc{AAWM} (Ours)}
  & \textbf{31.4}\stdv{4.0} & \textbf{17.6}\stdv{5.0} & \textbf{28.7}\stdv{6.6}
  & \textbf{27.3}\stdv{4.8} & \textbf{24.2}\stdv{5.8} & \textbf{4.8}\stdv{3.0}
  & \textbf{24.7}\stdv{2.6}
  & 25.7\stdv{2.5} & \textbf{10.7}\stdv{1.2} \\
\aawmBackboneRule
\multirow{3}{*}{Qwen2.5-7B-Instruct}
  & Base
  & 50.3\stdv{3.2} & 11.4\stdv{2.4} & 28.6\stdv{3.6}
  & 27.3\stdv{3.1} & 18.2\stdv{2.8} &  9.7\stdv{2.2}
  & 28.4\stdv{2.8}
  & 19.9\stdv{2.2} & 11.3\stdv{2.1} \\
  & \textsc{IWM}
  & 69.4\stdv{5.2} & 31.7\stdv{5.8} & 55.3\stdv{6.4}
  & 51.6\stdv{5.6} & 40.5\stdv{6.1} & \textbf{19.6}\stdv{4.6}
  & 49.6\stdv{4.8}
  & 27.4\stdv{5.3} &  9.0\stdv{3.5} \\
\rowcolor{gray!15}
  & \textbf{\textsc{AAWM} (Ours)}
  & \textbf{70.5}\stdv{5.6} & \textbf{35.7}\stdv{6.2} & \textbf{60.4}\stdv{6.8}
  & \textbf{55.6}\stdv{5.9} & \textbf{52.9}\stdv{6.4} & 14.7\stdv{5.0}
  & \textbf{53.4}\stdv{5.0}
  & \textbf{37.0}\stdv{2.9} & \textbf{16.3}\stdv{2.3} \\
\midrule
\aawmPanelRow{Imitation Learning then Reinforcement Learning initialized by World Modeling}
\midrule
\multirow{3}{*}{Qwen2.5-1.5B-Instruct}
  & Base
  & 82.1\stdv{7.3} & 61.5\stdv{2.7} & 81.9\stdv{7.5}
  & 85.1\stdv{6.3} & 58.9\stdv{1.6} & 59.0\stdv{7.7}
  & 74.0\stdv{4.6}
  & 71.1\stdv{4.7} & 59.4\stdv{2.7} \\
  & \textsc{IWM}
  & \textbf{94.4}\stdv{1.5} & 64.5\stdv{7.7} & 78.6\stdv{4.6}
  & 80.2\stdv{4.5} & 60.7\stdv{3.8} & 61.3\stdv{7.0}
  & 76.8\stdv{2.8}
  & 77.0\stdv{3.0} & 67.2\stdv{4.1} \\
\rowcolor{gray!15}
  & \textbf{\textsc{AAWM} (Ours)}
  & 93.3\stdv{4.6} & \textbf{72.0}\stdv{7.7} & \textbf{86.9}\stdv{5.2}
  & \textbf{88.7}\stdv{2.2} & \textbf{71.7}\stdv{3.1} & \textbf{71.9}\stdv{2.8}
  & \textbf{83.1}\stdv{1.7}
  & \textbf{81.5}\stdv{1.1} & \textbf{73.4}\stdv{3.6} \\
\aawmBackboneRule
\multirow{3}{*}{Qwen2.5-7B-Instruct}
  & Base
  & 93.0\stdv{4.2} & 66.7\stdv{3.7} & 92.8\stdv{6.0}
  & 91.2\stdv{1.7} & 79.3\stdv{5.4} & 68.6\stdv{2.4}
  & 84.6\stdv{1.3}
  & 80.8\stdv{4.9} & 70.6\stdv{4.8} \\
  & \textsc{IWM}
  & 92.6\stdv{2.2} & 76.3\stdv{2.2} & \textbf{94.4}\stdv{5.2}
  & 93.7\stdv{5.5} & 74.1\stdv{5.2} & 77.4\stdv{7.7}
  & 86.7\stdv{2.0}
  & 82.2\stdv{1.9} & 73.2\stdv{2.5} \\
\rowcolor{gray!15}
  & \textbf{\textsc{AAWM} (Ours)}
  & \textbf{95.2}\stdv{5.6} & \textbf{85.3}\stdv{2.9} & 93.1\stdv{2.5}
  & \textbf{94.1}\stdv{5.5} & \textbf{81.5}\stdv{7.6} & \textbf{83.4}\stdv{7.6}
  & \textbf{90.1}\stdv{2.0}
  & \textbf{84.2}\stdv{0.9} & \textbf{76.6}\stdv{2.1} \\
\bottomrule
\end{tabular}
\endgroup
}
\caption{Main results on ALFWorld and WebShop.
In trained panels, Base denotes no world modeling initialization, \textsc{IWM} denotes next-observation world modeling, and \textsc{AAWM} denotes Agent-Authored World Modeling.
Trained rows report mean$\pm$std over three seeds; bold marks the best within each backbone and condition.}
\label{tab:main}
\end{table*}

\Cref{tab:main} shows that \textsc{AAWM} consistently improves over the world-modeling baseline on both agentic benchmarks and at both backbone scales.
Although closed-source prompting models remain competitive, with Gemini-2.5-Pro~\citep{geminiteam2025gemini25} and GPT-4o~\citep{openai2024gpt4osystemcard} reaching 60.3\% and 48.0\% success on ALFWorld and 35.9\% and 23.7\% on WebShop, post-training narrows and then reverses this gap for the Qwen-family models.
Among the trained methods, \textsc{AAWM} achieves the strongest overall performance.

\paragraph{\textsc{AAWM} provides a stronger initialization than next observation prediction.}
At the 1.5B scale, Base reaches 74.0\% success on ALFWorld, while
\textsc{IWM} improves it to 76.8\% and \textsc{AAWM} further improves it
to 83.1\%.
WebShop shows the same ordering, with success increasing from 59.4\%
(Base) to 67.2\% (\textsc{IWM}) and 73.4\% (\textsc{AAWM}).
At the 7B scale, \textsc{AAWM} again achieves the strongest aggregate
performance, reaching 90.1\% on ALFWorld and 76.6\% on WebShop.

\paragraph{\textsc{AAWM} consistently outperforms on the harder tasks.}
The ALFWorld breakdown shows that the aggregate gain is concentrated on
the more difficult categories.
Pick and Clean are easier for most methods, and they are the only
categories where \textsc{IWM} slightly exceeds \textsc{AAWM}: Pick at
1.5B and Clean at 7B.
On the harder categories, the ordering reverses and the margins become
larger.
At 1.5B, \textsc{AAWM} improves over \textsc{IWM} by 11.0 points on
Cool, 10.6 on Pick2, and 7.5 on Look.
At 7B, the gains are 9.0 points on Look, 7.4 on Cool, and 6.0 on Pick2.
These tasks require reasoning over multi-step preconditions and
persistent object states, which are difficult to recover from a single
observation.
This is where Self-Probing and cross-trajectory retrieval are expected
to be most useful, because they expose decision-relevant dynamics beyond
the immediate observation.

\paragraph{\textsc{AAWM} produces a more decision-oriented initialization.}
A further question is whether this advantage comes from the world modeling targets themselves or from their interaction with the imitation learning stage.
To isolate this, we remove the imitation learning stage so that world modeling alone carries the full initialization budget before GRPO (Appendix~\ref{app:direct-rl}).
\textsc{IWM} then drops sharply on ALFWorld, while \textsc{AAWM} stays close to GRPO from the backbone and still improves WebShop.
Since both methods share the same Qwen3-30B-A3B-Instruct-2507 construction model on WebShop, this comparison controls for constructor capacity, indicating that the gap is not explained by construction-model scale.
\textsc{AAWM} depends far less on action imitation because its targets already describe the dynamics the policy needs when selecting actions.

\subsection{Ablation Studies}
\label{sec:exp:ablation}

\begin{table}[!h]\small
\centering
\resizebox{.99\columnwidth}{!}{%
\begingroup
\setlength{\tabcolsep}{5.5pt}
\begin{tabular}{@{}l c c@{}}
\toprule
Method & ALFWorld & WebShop \\
\midrule
\multicolumn{3}{@{}c@{}}{\textbf{Imitation Learning initialized by World Modeling}} \\
\midrule
Base                                  & 12.0\stdv{2.6}          &  4.2\stdv{1.2}          \\
\rowcolor{gray!15}
\textbf{\textsc{AAWM} (Ours)}         & \textbf{24.7}\stdv{2.6} & \textbf{10.7}\stdv{1.2} \\
\quad w/o Transition Retrieval        & 19.3\stdv{2.4}          &  7.5\stdv{1.6}          \\
\quad w/o Self-Probing                & 18.0\stdv{1.6}          &  7.0\stdv{3.1}          \\
CoT-Only                              & 14.8\stdv{2.1}          &  5.2\stdv{1.6}          \\
Hindsight Reflection                  & 20.8\stdv{1.6}          &  6.2\stdv{1.4}          \\
\bottomrule
\end{tabular}
\endgroup
}
\caption{Ablation at Qwen2.5-1.5B-Instruct.
All rows use the same lightweight imitation learning stage and are evaluated without GRPO.
Base denotes imitation learning without world modeling initialization.
ALFWorld reports aggregate success rate, and WebShop reports success rate.
All cells report mean$\pm$std over three seeds.}
\label{tab:ablation}
\end{table}

\Cref{tab:ablation} separates two questions about the source of \textsc{AAWM}'s gains.
The component ablations remove \textit{Self-Probing} or \textit{Transition Retrieval}, while the target comparisons replace \textsc{AAWM} with generic reasoning or hindsight supervision.
All rows use the 1.5B backbone and share the same imitation learning stage.
Full \textsc{AAWM} improves ALFWorld from 12.0\% to 24.7\% and WebShop from 4.2\% to 10.7\% over Base.

\paragraph{Retrieved evidence supplies dynamics beyond the current state.}
Removing Transition Retrieval lowers the results to 19.3\% on ALFWorld and 7.5\% on WebShop.
Without retrieved evidence, \textit{Dynamics Synthesis} sees only the immediate transitions at the current state, which often do not contain enough information to resolve the policy's open questions.

\paragraph{Self-Probing tells synthesis which dynamics to address.}
Removing Self-Probing lowers the results further to 18.0\% and 7.0\%.
Without the policy's confirmed patterns and open questions, the synthesis model has no explicit signal about what the target should cover.
The two components together outperform either alone, showing that both contribute to shifting the training target from environment-driven content to decision-driven content.

\paragraph{AAWM outperforms alternative supervision targets.}
We next test whether generic reasoning or reflection-style supervision can reproduce the benefit of \textsc{AAWM}'s target construction.
CoT-Only uses Qwen3-30B-A3B-Thinking-2507 to generate reasoning from $(o_t,a_t)$ without transition evidence, while Hindsight Reflection uses the same transition evidence and synthesis model as \textsc{AAWM} to write a hindsight lesson directly.
CoT-Only gives a modest improvement over Base, and Hindsight Reflection is stronger, but both remain below full \textsc{AAWM} on the two environments.
A separate State-Change control uses the same construction model to describe only the task-relevant difference between $o_t$ and $o_{t+1}$ rather than reconstructing the full next observation.
This single-run control reaches 9.4\% on ALFWorld and 5.7\% on WebShop (Appendix~\ref{app:state-change}).
Appendix~\ref{app:target-comparison} provides direct qualitative comparisons between next-observation targets and \textsc{AAWM} targets for matched decision contexts.
Overall, the advantage of \textsc{AAWM} is not explained by generic reasoning, hindsight supervision, or transition abstraction alone.

\subsection{Evaluation on More Environments}
\label{sec:exp:agentgym}

The main experiments use two environments with GRPO.
To test whether decision-oriented targets remain beneficial across a broader set of environments under supervised training alone, we evaluate on four environments from AgentGym~\citep{xi2025agentgym}.
All methods train Qwen2.5-1.5B-Instruct on the same mixture of AgentTraj-L trajectories from TextCraft, SciWorld, WebShop, and ALFWorld, and use the same imitation learning stage.
We also include \textsc{IWM}-Summary, a variant that uses the same Qwen3-30B-A3B model as \textsc{AAWM} to summarize observations before prediction, which controls for the capacity of the construction model.
Data counts and evaluation settings are in Appendix~\ref{app:agentgym-details}.

\begin{table*}[!t]
\centering\small
\resizebox{.99\textwidth}{!}{%
\begingroup
\setlength{\tabcolsep}{12pt}
\begin{tabular}{@{}l ccccc@{}}
\toprule
Method & TextCraft & SciWorld & WebShop & ALFWorld & All \\
\midrule
\multicolumn{6}{c}{\textbf{Prompting}} \\
\midrule
Qwen2.5
  & 18.3\stdv{1.5} & 11.7\stdv{3.1} & 21.3\stdv{0.8} &  3.5\stdv{1.8} & 13.0\stdv{0.5} \\
\midrule
\multicolumn{6}{c}{\textbf{Imitation Learning initialized by World Modeling}} \\
\midrule
Base
  & 31.7\stdv{1.5} & 70.2\stdv{2.8} & 94.5\stdv{0.9} & 33.3\stdv{2.0} & 61.1\stdv{1.8} \\
\textsc{IWM}
  & 31.3\stdv{1.5} & 79.8\stdv{2.8} & 83.0\stdv{2.5} & 30.2\stdv{2.3} & 59.6\stdv{2.4} \\
\textsc{IWM}-Summary
  & 32.3\stdv{2.1} & 84.2\stdv{2.8} & 94.5\stdv{2.3} & 20.0\stdv{2.2} & 61.4\stdv{2.0} \\
\rowcolor{gray!15}
\textbf{\textsc{AAWM} (Ours)}
  & \textbf{35.7}\stdv{1.2} & \textbf{91.2}\stdv{1.8} & \textbf{97.5}\stdv{0.5} & \textbf{36.0}\stdv{1.8} & \textbf{69.3}\stdv{1.3} \\
\bottomrule
\end{tabular}
\endgroup
}
\caption{Results on four AgentGym environments at Qwen2.5-1.5B-Instruct.
All trained rows share the same AgentTraj-L trajectory mixture and imitation learning stage.
All is weighted by the number of evaluation tasks per environment.
Cells report success rate (\%), mean$\pm$std over three seeds.}
\label{tab:agentgym}
\end{table*}

\Cref{tab:agentgym} reports success rate in each environment and the weighted aggregate All.
Base reaches an All of 61.1\%.
The two next observation variants do not give a consistent improvement.
\textsc{IWM} falls to 59.6\%, because predicting raw observations helps SciWorld but hurts WebShop and ALFWorld.
\textsc{IWM}-Summary recovers WebShop and improves SciWorld for an All of 61.4\%, yet it drops sharply on ALFWorld to 20.0\%.
\textsc{AAWM} instead reaches the best result in all four environments, with an All of 69.3\% and the largest single gain of 21.0 points on SciWorld.
It is the only world modeling initialization that improves over Base in every environment.
Because all methods share the backbone, the trajectory mixture, and the imitation learning stage, this pattern across the four evaluated environment families supports targets shaped by the policy's decision needs as a more reliable learning signal than targets derived from the next observation.

\section{Analysis}
\label{sec:analysis}

The experiments establish the performance gains of \textsc{AAWM} over next-observation prediction.
We next examine how its initialization affects early GRPO behavior and whether \textit{Self-Probing} and \textit{Dynamics Synthesis} form a useful, evidence-grounded target construction process.
We then study the number of propositions and control for supervision budget and synthesizer scale.

\begin{figure}[t]
\centering
\includegraphics[width=\columnwidth]{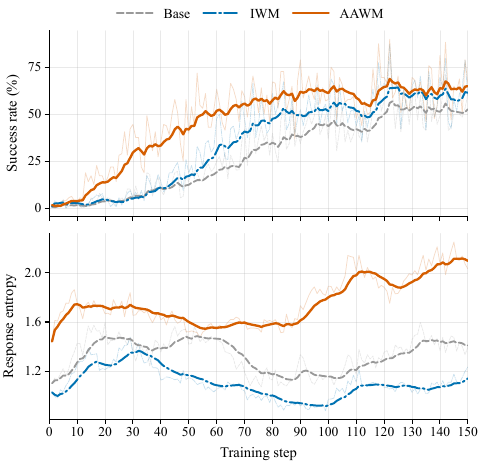}
\caption{GRPO training dynamics on WebShop under the imitation learning then reinforcement learning setting initialized by world modeling.
The backbone is Qwen2.5-1.5B-Instruct.
The top panel shows success rate, and the bottom panel shows response entropy.
Bold curves show time weighted EMA smoothing with coefficient 0.95.
\textsc{AAWM} converges faster while maintaining higher response entropy during training.}
\label{fig:webshop-dynamics}
\end{figure}

\paragraph{Training dynamics.}
A policy that converges faster could be collapsing to a narrow output distribution or exploring more effectively.
We first track success rate and response entropy during GRPO, alongside response length and rollout content. We compute response entropy as the token-level entropy of the policy's next-token distribution, averaged over generated response tokens.
\Cref{fig:webshop-dynamics} shows the WebShop dynamics, and Appendix~\ref{app:alfworld-dynamics} reports the matching ALFWorld curves.
The success curves show that \textsc{AAWM} learns faster.
On WebShop all three methods stay low during the first ten steps, after which \textsc{AAWM} rises much more rapidly than Base and \textsc{IWM} and keeps the highest success rate for most of training, and ALFWorld follows the same trend.
The entropy curves show that this faster convergence does not coincide with an early collapse in the response distribution.

To complement the entropy curves with direct behavioral evidence, we pool equal-budget rollouts from GRPO steps 0 and 50.
We measure how broadly the policies visit environment states, how many distinct executable actions they take, and how many distinct state-action pairs occur in successful rollouts.
\Cref{tab:behavioral-exploration} reports these three aspects of early exploration.

\begin{table}[!t]\small
\centering
\resizebox{.99\columnwidth}{!}{%
\begin{tabular}{@{}llrrr@{}}
\toprule
Environment & Method & States & Valid actions & Success pairs \\
\midrule
\multirow{2}{*}{ALFWorld}
  & \textsc{IWM}  & \textbf{932} & \textbf{375} & 629 \\
  & \textsc{AAWM} & 887 & 342 & \textbf{683} \\
\midrule
\multirow{2}{*}{WebShop}
  & \textsc{IWM}  & 376 & 181 & 141 \\
  & \textsc{AAWM} & \textbf{468} & \textbf{218} & \textbf{199} \\
\bottomrule
\end{tabular}}
\caption{Early GRPO behavior under equal rollout budgets, pooling steps 0 and 50.
States, Valid actions, and Success pairs count distinct visited states, distinct executable actions taken, and distinct state-action pairs in successful rollouts, respectively.}
\label{tab:behavioral-exploration}
\end{table}

On ALFWorld, \textsc{AAWM} visits slightly fewer raw states and actions but reaches more successful state-action pairs.
On WebShop, it improves both raw coverage and successful behavior diversity.
Its outputs are also longer on average, with 167.6 tokens on WebShop and 98.7 on ALFWorld against 121.4 and 57.8 for \textsc{IWM}, and the extra tokens appear mainly in the reasoning trace.
Representative rollouts in Appendix~\ref{app:rollout} confirm that the \textsc{AAWM} initialized policy reasons about environment mechanisms before acting, including object persistence, action preconditions, and product attributes.
Together, the behavioral measures and training dynamics support more success-oriented exploration after \textsc{AAWM} initialization, rather than uniformly greater raw coverage.

\paragraph{Alignment of Self-Probing and Synthesis.}
We evaluate whether the two stages of target construction operate as intended.
A useful process should identify propositions that matter for action choice, resolve open questions when the evidence permits, correct mistaken beliefs, and keep synthesized claims grounded in the supplied transitions.
For each environment, we randomly sample 100 confirmed patterns and 100 open questions.
Three independent LLM judges, \mbox{Qwen3.6-Flash}, DeepSeek-V4-Flash~\citep{deepseekai2026deepseekv4}, and GLM-5~\citep{glm5team2026glm5vibecodingagentic}, evaluate each example, and their majority vote gives the model score in \Cref{tab:judge}.
We also obtain human labels for 10 randomly sampled examples of each proposition type in each environment.
\textit{Decision helpfulness} measures whether a proposition could inform action choice.
\textit{Question resolution} and \textit{pattern correction} measure how synthesis handles open questions and mistaken beliefs.
\textit{Factual correctness} measures whether the synthesized claims are supported by the supplied transitions.

\begin{table}[!t]\small
\centering
\resizebox{.99\columnwidth}{!}{%
\begin{tabular}{@{}lccc@{}}
\toprule
Criterion & Majority & Human & Agreement / $\kappa$ \\
\midrule
Decision helpfulness & 96.5\% & 95\% & 95\% / .643 \\
Question resolution & 85\% & 80\% & 90\% / .737 \\
Pattern correction & 17\% & 30\% & 70\% / .348 \\
Factual correctness & 80\% & 70\% & 80\% / .524 \\
\bottomrule
\end{tabular}}
\caption{Independent evaluation of \textit{Self-Probing} and \textit{Dynamics Synthesis}.
Majority is the positive rate under majority vote from the three LLM judges.
Agreement and Cohen's $\kappa$ compare the majority vote with human labels.}
\label{tab:judge}
\end{table}

Decision helpfulness and question resolution are both high, indicating that Self-Probing focuses on information useful for action choice and that synthesis usually answers the resulting open questions.
Pattern correction is lower, though synthesis still revises some confirmed patterns.
This rate is naturally lower because a positive case requires the original pattern to be mistaken and the synthesis output to correct it.
Factual correctness further shows that most synthesized claims are supported by the transition evidence.
Together, these results support the decision relevance and evidence grounding of the constructed targets.
Representative triples are in Appendix~\ref{app:probing-pairs}.

\paragraph{Sensitivity to the number of propositions.}
We vary the number of confirmed patterns and open questions $n$ produced at each state (Appendix~\ref{app:n-sensitivity}).
Increasing $n$ from 1 to 3 clearly improves both environments, suggesting that a single proposition may not cover the dynamics needed for action selection, while moving from 3 to 5 adds little.
We therefore use $n=3$ in all main experiments as a compact setting that captures most of the benefit.

\paragraph{Supervision and synthesizer controls.}
Finally, we test whether the gains can be attributed to a larger supervision budget or reliance on the 30B synthesis model.
Appendix~\ref{app:additional-controls} reports target token counts, matched-budget comparisons, construction cost, and synthesizer scale.
Although each \textsc{AAWM} target is longer on average, using one target per state gives 46\% fewer assistant target tokens on ALFWorld and 71\% fewer on WebShop over two world modeling SFT epochs.
\textsc{AAWM} remains stronger than \textsc{IWM} at every matched token budget, and the advantage remains when the 7B policy itself serves as the synthesizer.
These controls show that the gains do not depend on more supervised tokens or a separate 30B construction model.

\section{Conclusion}
\label{sec:conclusion}

World modeling for language agents has typically predicted the next environment output, but prediction accuracy and decision quality serve different goals.
AAWM addresses this gap by letting the policy state what it understands and what remains uncertain, then constructing training targets from these statements.
Across model scales, training settings, and environment families, AAWM consistently outperforms next-observation world modeling.
A natural extension is to repeat \textit{Self-Probing} after each training round so that targets evolve with the policy's understanding.
More broadly, these results suggest that world modeling targets should be evaluated by the decisions they help the policy make.

\section*{Limitations}
\label{sec:limitations}

\textsc{AAWM} runs \textit{Self-Probing} once during data construction, so the training target reflects the policy's beliefs at that point rather than its beliefs as training proceeds.
Repeating \textit{Self-Probing} after each training round could turn the target into a curriculum that follows the policy's changing understanding, at the cost of additional construction compute, and we leave this to future work.

Our experiments also cover text-only environments.
Extending \textsc{AAWM} to multimodal settings, where a vision language policy must model dynamics from both images and text, is a natural next step that we have not yet explored.

All trained backbones in our experiments come from the Qwen model family.
We report three-seed means for the main comparisons and mark single-run controls explicitly.
A broader statistical study across more model families remains future work.

\section*{Ethical Considerations}

This work studies world modeling for language agents in sandboxed benchmark environments. 
Because the proposed method improves sequential decision-making ability, similar techniques could potentially be misused in open-ended web, GUI, or tool-use settings where autonomous actions may have external consequences. 
 
Our experiments do not deploy agents in real-world systems, and practical deployment would require additional safeguards, monitoring, and validation of generated targets.

\ifanonymousawardcopy
\else
\section*{Acknowledgments}
This research is supported by the Big Data Computing Center of Southeast University.
\fi

\bibliography{references}

\clearpage
\appendix
\raggedbottom 
\section{Experimental Details}
\label{app:exp-details}

\subsection{Main Experiment Details}
\label{app:main-exp-details}

\noindent\textbf{Trajectory selection.}
AgentTraj-L is a collection of high-quality agent trajectories released with AgentGym~\citep{xi2025agentgym}.
For the main ALFWorld and WebShop experiments, we select complete trajectories whose action sequences can be executed from start to finish in the GiGPO open-source evaluation code and reach a successful outcome.
The selected set $\mathcal{D}$ contains 1{,}655 trajectories with 19{,}039 action steps for ALFWorld, and 1{,}027 trajectories with 4{,}971 action steps for WebShop.

\noindent\textbf{Transition collection.}
We build the transition bank from every action step in $\mathcal{D}$.
At each step, we execute the action in the trajectory and three additional actions from the same state, so each step contributes exactly four one-step transitions.
For ALFWorld, the three additional actions are sampled uniformly without replacement from the admissible action set after removing the trajectory action and \texttt{help}.
For WebShop, two additional actions are sampled from the matched policy backbone with temperature 0.8, and one additional action is sampled uniformly from the admissible action set after removing the trajectory action.
Sampled actions are canonicalized and deduplicated.
If a policy sampled action cannot be executed or duplicates an existing action, we replace it with an unused admissible action.
This procedure produces exactly three additional actions for every step.
The final transition bank contains 76{,}156 one-step transitions for ALFWorld and 19{,}884 one-step transitions for WebShop.

\noindent\textbf{Training and evaluation separation.}
All trajectories in $\mathcal{D}$ and all records in $\mathcal{T}$ come from the training split.
Evaluation trajectories do not enter world modeling data construction, and retrieval excludes records from the same episode as the query.
We find no identical task or goal text between the training and evaluation splits.
As an additional check within the training data, we compare each query goal with the goal of every retrieved source trajectory.
\Cref{tab:retrieval-overlap} shows that exact and near nonexact matches are rare.

\begin{table}[!t]\small
\centering
\begin{tabular*}{.99\columnwidth}{@{\extracolsep{\fill}}lcc@{}}
\toprule
Environment & Exact goal & Near nonexact goal \\
\midrule
ALFWorld & 0.9\% & 0.0\% \\
WebShop & 0.0\% & 0.3\% \\
\bottomrule
\end{tabular*}
\caption{Goal overlap between training queries and retrieved source trajectories.
Near nonexact matches have token set Jaccard similarity $0.95 \leq J < 1$.}
\label{tab:retrieval-overlap}
\end{table}

\noindent\textbf{Supervised fine-tuning settings.}
World model fine-tuning and imitation learning both use full-parameter supervised fine-tuning.
The learning rate is $1\times 10^{-5}$.
The effective batch size is 16 for ALFWorld and 8 for WebShop.
The world modeling stage uses two epochs for both \textsc{IWM} and \textsc{AAWM}.
The imitation learning stage uses one epoch.
The imitation learning subset contains 32 trajectories from $\mathcal{D}$, with 429 action steps for ALFWorld and 158 action steps for WebShop.
For each action step, the model is supervised to generate the action content in the format \texttt{<action>...</action>}.

\noindent\textbf{GRPO settings.}
We use GRPO for all reinforcement learning experiments and follow the public implementation of \citet{feng2025gigpo}.
For ALFWorld, the maximum prompt length is 2048 tokens, the maximum response length is 512 tokens, and each episode allows up to 50 environment steps.
For WebShop, the maximum prompt length is 4096 tokens, the maximum response length is 512 tokens, and each episode allows up to 15 environment steps.
Both environments use a rule-based reward of 10 for success and 0 for failure, with a penalty of $-0.1$ for invalid actions.
The actor learning rate is $1\times 10^{-6}$.
For group-based RL, the group size is 8 and each rollout samples 16 groups, giving 128 environments in total.
The rollout temperature is 1.0, the validation temperature is 0.4, and the KL coefficient is 0.01.
The mini batch size is 256 for ALFWorld and 64 for WebShop.

\subsection{Multi-Environment Evaluation Details}
\label{app:agentgym-details}

The multi-environment experiment uses the AgentGym evaluation code and is separate from the main ALFWorld and WebShop setting.
It trains on the full AgentTraj-L train split for TextCraft, SciWorld, WebShop, and ALFWorld, and evaluates on the full AgentTraj-L eval split for each environment.
No additional branch transitions are collected in this experiment, because the mixed train split already provides broad transition coverage across environments.
For \textsc{AAWM}, the immediate transition set $\mathcal{I}_t$ contains only the original one-step transition at the corresponding step.

The training data contain 2{,}420 trajectories for ALFWorld, 3{,}930 for WebShop, 374 for TextCraft, and 2{,}120 for SciWorld, for a total of 8{,}844 trajectories.
Evaluation uses 200 tasks each for ALFWorld, WebShop, and SciWorld, and 100 tasks for TextCraft.
The backbone is Qwen2.5-1.5B-Instruct.
All supervised runs use full fine-tuning with learning rate $1\times 10^{-5}$ and effective batch size 32.
\textsc{IWM}, \textsc{IWM}-Summary, and \textsc{AAWM} use two epochs of world model fine-tuning followed by one epoch of imitation learning.
Base only uses one epoch of imitation learning without world model fine-tuning.
The \textsc{IWM}-Summary baseline summarizes observations in all four environments using Qwen3-30B-A3B-Instruct-2507 before next-observation prediction.
Evaluation uses greedy decoding with temperature 0.0.
The maximum numbers of interaction rounds are 20 for TextCraft, 30 for SciWorld, 6 for WebShop, and 30 for ALFWorld.

\subsection{Sensitivity to Self-Probing Propositions}
\label{app:n-sensitivity}

\Cref{tab:n-sensitivity} varies the number of confirmed patterns and open questions $n$ produced at each state.
Increasing $n$ from 1 to 3 gives a clear improvement on both environments, showing that a single proposition is insufficient to cover the dynamics needed for action selection.
Increasing $n$ from 3 to 5 yields smaller additional gains.
We therefore use $n=3$ in all main experiments as a compact setting that captures most of the benefit.

\begin{table}[!t]\small
\centering
\begin{tabular*}{.99\columnwidth}{@{\extracolsep{\fill}} c c c @{}}
\toprule
$n$ & ALFWorld & WebShop \\
\midrule
1 & 10.7\stdv{2.2}  &  7.8\stdv{0.8} \\
3 & 24.7\stdv{2.6}  & 10.7\stdv{1.2} \\
5 & 26.0\stdv{1.2}  & 11.7\stdv{0.8} \\
\bottomrule
\end{tabular*}
\caption{Sensitivity to the number of Self-Probing propositions $n$ at Qwen2.5-1.5B-Instruct.
Each setting uses the same world model fine-tuning and lightweight imitation learning protocol and is evaluated without GRPO.
ALFWorld reports aggregate success rate, and WebShop reports success rate.
Cells report mean$\pm$std over three seeds.}
\label{tab:n-sensitivity}
\end{table}

\section{Additional Control Analyses}
\label{app:additional-controls}
\FloatBarrier

\subsection{State-Change Target}
\label{app:state-change}

State-Change uses the same construction model and training protocol as the main 1.5B experiments, but its target describes only the task-relevant difference between $o_t$ and $o_{t+1}$.
In a single run, it reaches 9.4\% success on ALFWorld and 5.7\% on WebShop.
This control remains below \textsc{AAWM} on both environments, suggesting that transition abstraction alone does not reproduce decision-oriented target construction.

\begin{table}[H]\small
\centering
\begin{tabular*}{.99\columnwidth}{@{\extracolsep{\fill}} l c c @{}}
\toprule
Method & ALFWorld & WebShop \\
\midrule
State-Change & 9.4 & 5.7 \\
\bottomrule
\end{tabular*}
\caption{Single-run results for the State-Change target at Qwen2.5-1.5B-Instruct.
Both environments report success rate.}
\label{tab:state-change}
\end{table}

\subsection{Supervision Budgets}

\Cref{tab:target-tokens} reports the assistant target tokens used during two epochs of world modeling SFT.
Each \textsc{AAWM} target is longer on average, but \textsc{AAWM} produces one target per state while \textsc{IWM} produces one target for each of four transitions.
As a result, \textsc{AAWM} uses 46\% fewer target tokens on ALFWorld and 71\% fewer on WebShop.

\begin{table}[H]\small
\centering
\resizebox{.99\columnwidth}{!}{%
\begin{tabular}{@{}llrrr@{}}
\toprule
Environment & Method & Targets / state & Avg. tokens & Total tokens \\
\midrule
\multirow{2}{*}{ALFWorld}
  & \textsc{IWM}  & 4 & 44.5 & 6.780M \\
  & \textsc{AAWM} & 1 & 96.3 & 3.665M \\
\midrule
\multirow{2}{*}{WebShop}
  & \textsc{IWM}  & 4 & 98.8 & 3.928M \\
  & \textsc{AAWM} & 1 & 113.2 & 1.125M \\
\bottomrule
\end{tabular}}
\caption{Assistant target tokens over two world modeling SFT epochs.}
\label{tab:target-tokens}
\end{table}

We also train both methods at the original \textsc{AAWM} budget $B_A$, the mean budget $B_M$, and the original \textsc{IWM} budget $B_I$.
\Cref{tab:matched-budgets} shows that \textsc{AAWM} remains stronger at every matched point.
The benefit therefore does not come from receiving more supervised target tokens.

\begin{table}[H]\small
\centering
\begin{tabular*}{.99\columnwidth}{@{\extracolsep{\fill}}llrrr@{}}
\toprule
Environment & Method & $B_A$ & $B_M$ & $B_I$ \\
\midrule
\multirow{2}{*}{ALFWorld}
  & \textsc{IWM}  & 3.1 & 5.4 & 3.1 \\
  & \textsc{AAWM} & \textbf{24.7} & \textbf{19.6} & \textbf{25.0} \\
\midrule
\multirow{2}{*}{WebShop}
  & \textsc{IWM}  & 1.6 & 2.3 & 2.1 \\
  & \textsc{AAWM} & \textbf{10.7} & \textbf{9.4} & \textbf{4.7} \\
\bottomrule
\end{tabular*}
\caption{Success rates under matched assistant target token budgets.
The budgets $B_A$, $B_M$, and $B_I$ are 3.67M, 5.22M, and 6.78M tokens on ALFWorld and 1.13M, 2.53M, and 3.93M tokens on WebShop.}
\label{tab:matched-budgets}
\end{table}

\subsection{Construction Cost and Synthesizer Scale}

Target construction is performed once before downstream training.
\Cref{tab:construction-cost} shows that most of its token use comes from Self-Probing and Dynamics Synthesis, while retrieval takes a small fraction of runtime.
Each target uses three confirmed patterns and three open questions, with three retrieved records for each proposition.
It therefore receives 18 retrieved records by design.

\begin{table*}[!t]
\centering\small
\resizebox{.99\textwidth}{!}{%
\begingroup
\setlength{\tabcolsep}{12pt}
\begin{tabular}{@{}lrrrr@{}}
\toprule
Environment & Self-Probing input / output & Synthesis input / output & Retrieval time & Branch output \\
\midrule
ALFWorld & 12.70M / 2.05M & 34.85M / 1.83M & 0.3\% & 0 \\
WebShop  &  4.70M / 0.45M & 22.55M / 0.56M & 0.1\% & 0.175M \\
\bottomrule
\end{tabular}
\endgroup
}
\caption{Offline construction accounting at the 1.5B policy scale.
All token values are totals over the construction set.
Retrieval time is its share of total construction runtime.
Branch output counts tokens generated for alternative actions: ALFWorld samples all alternatives uniformly without an LLM call, whereas WebShop samples two of the three from the policy backbone.}
\label{tab:construction-cost}
\end{table*}

We further use Qwen2.5-7B-Instruct as both the policy and the \textsc{AAWM} synthesizer.
The same 7B model produces the WebShop summaries for \textsc{IWM}, while ALFWorld \textsc{IWM} continues to use raw observations.
\Cref{tab:synthesizer-scale} shows that \textsc{AAWM} retains its advantage with this matched 7B constructor.
The smaller synthesizer nearly preserves ALFWorld performance and gives a smaller result on WebShop, where synthesis must integrate more product information.

\begin{table}[!t]\small
\centering
\begin{tabular*}{.99\columnwidth}{@{\extracolsep{\fill}}lccc@{}}
\toprule
Method & Constructor & ALFWorld & WebShop \\
\midrule
\textsc{IWM} & raw / 7B summary & 49.6\stdv{4.8} & 9.1\stdv{3.5} \\
\textsc{AAWM} & 7B & 52.3\stdv{2.1} & 13.0\stdv{1.6} \\
\textsc{AAWM} & 30B & \textbf{53.4}\stdv{5.0} & \textbf{16.3}\stdv{2.3} \\
\bottomrule
\end{tabular*}
\caption{Synthesizer scale at the Qwen2.5-7B-Instruct policy scale.
Cells report mean$\pm$std over three seeds.}
\label{tab:synthesizer-scale}
\end{table}

\FloatBarrier
\section{Direct Comparison of World Modeling Targets}
\label{app:target-comparison}

The following two examples compare the supervision generated by next-observation world modeling and \textsc{AAWM} for matched decision contexts.

\begin{casebox}{ALFWorld case}
\textbf{Task.}\;
heat some cup and put it in cabinet.

\smallskip
\textbf{Current observation.}\;
You arrive at microwave~1. The microwave~1 is closed.

\tcbline
\textbf{\textsc{IWM} target.}\;
\textit{(given the action ``\texttt{heat cup 1 with microwave 1}'')}\\
You heat the cup~1 using the microwave~1.

\tcbline
\textbf{\textsc{AAWM} target.}\;
The environment allows heating an item with the microwave without requiring the microwave to be open or the item to be placed inside; the heating action automatically processes the item while it is in hand, and the item remains in hand afterward. Consequently, the microwave's state does not affect the heating action, and no retrieval from the microwave is needed after heating.

\tcbline
\textbf{Interpretation.}\;
The \textsc{IWM} target states only the immediate outcome and misses the unusual action preconditions and postconditions. The \textsc{AAWM} target captures the decision-critical dynamics: the closed microwave does not block heating, and the cup remains in hand, so the agent can proceed directly to the cabinet without opening or retrieving it.
\end{casebox}

\begin{casebox}{WebShop variant-selection case}
\textbf{Task.}\;
Find me hand wash men's sleep \& lounge with long sleeve, elastic waistband for daily wear with color: multi 9, and size: medium, and price lower than 80.00 dollars.

\smallskip
\textbf{Current observation.}\;
\texttt{Instruction: [SEP] Find me hand wash men's sleep \& lounge with long sleeve, elastic waistband for daily wear with color: multi 9, and size: medium, and price lower than 80.00 dollars [SEP] Back to Search [SEP] < Prev [SEP] size [SEP] small [SEP] medium [SEP] large [SEP] x-large [SEP] xx-large [SEP] color [SEP] multi 1 [SEP] multi 10 [SEP] multi 2 [SEP] multi 3 [SEP] multi 4 [SEP] multi 5 [SEP] multi 6 [SEP] multi 7 [SEP] multi 8 [SEP] multi 9 [SEP] InterestPrint Deer and Stars Men's Loungewear Pajama Sets - Long Sleeve Tee and Jogger Pant [SEP] Price: \$43.59 [SEP] Rating: N.A. [SEP] Description [SEP] Features [SEP] Reviews [SEP] Buy Now}

\tcbline
\textbf{\textsc{IWM}-Summary target.}\;
\textit{(given the action ``\texttt{click[medium]}'')}\\
This is a product search result page displaying a men's loungewear item matching the specified criteria. The active filters show size ``medium'' and color ``multi 9'' selected, with the featured product being ``InterestPrint Deer and Stars Men's Loungewear Pajama Sets'' priced at \$43.59. Key visible elements include the product name, price, and navigation options like ``Back to Search'' and ``Prev'', alongside the structured size and color filter lists. The page confirms the search parameters are applied and a relevant item is listed within the price range.

\tcbline
\textbf{\textsc{AAWM} target.}\;
The environment requires explicit selection of size and color on the product page to capture the desired specifications for purchase. The search query's filters do not automatically apply to the product's variant options; instead, the system only records selections made via the product page's filters. Selecting size ``medium'' and color ``multi 9'' before purchasing ensures the correct variant is purchased.

\tcbline
\textbf{Interpretation.}\;
The \textsc{IWM}-Summary target describes the next page and even suggests that both requested options are already applied, but it does not expose the selection mechanism. The \textsc{AAWM} target captures the key dynamics: query attributes do not set product variants.
\end{casebox}

\section{Reinforcement Learning without Imitation Learning}
\label{app:direct-rl}

We isolate whether world modeling initialization remains useful when GRPO starts without the lightweight imitation learning stage.
\Cref{tab:direct-rl} compares direct GRPO from the backbone with GRPO from the two world modeling initializations at Qwen2.5-1.5B-Instruct.

Without imitation learning, \textsc{IWM} reaches only 13.2\% on ALFWorld and 54.2\% on WebShop.
This is far below direct GRPO from the backbone on ALFWorld and also lower on WebShop.
By contrast, \textsc{AAWM} reaches 70.6\% on ALFWorld and 63.8\% on WebShop.
It therefore preserves the direct GRPO performance on ALFWorld and improves WebShop over the base row.
This shows that \textsc{AAWM} relies less on imitation learning than next observation prediction.
A likely reason is that \textsc{AAWM} trains on decision-oriented dynamics, so its world modeling target is closer to the information used by the policy during reinforcement learning.

\begin{table}[!t]\small
\centering
\setlength{\tabcolsep}{8pt}
\begin{tabular}{@{}l c c@{}}
\toprule
Method & ALFWorld & WebShop \\
\midrule
\multicolumn{3}{@{}c}{\textbf{Reinforcement Learning initialized by World Modeling}} \\
\midrule
Base
  & \textbf{71.6}\stdv{2.8} & 58.6\stdv{3.8} \\
\textsc{IWM}
  & 13.2\stdv{3.2} & 54.2\stdv{2.7} \\
\rowcolor{gray!15}
\textbf{\textsc{AAWM} (Ours)}
  & 70.6\stdv{2.7} & \textbf{63.8}\stdv{2.7} \\
\bottomrule
\end{tabular}
\caption{Direct reinforcement learning at Qwen2.5-1.5B-Instruct.
All rows start GRPO without the lightweight imitation learning stage.
Base denotes GRPO directly from the backbone.
\textsc{IWM} and \textsc{AAWM} denote GRPO from the corresponding world modeling initialization.
ALFWorld reports aggregate success rate, and WebShop reports success rate.
Cells report mean$\pm$std over three seeds.}
\label{tab:direct-rl}
\end{table}

\section{Additional Training Dynamics}
\label{app:alfworld-dynamics}

\begin{figure}[t]
\centering
\includegraphics[width=\columnwidth]{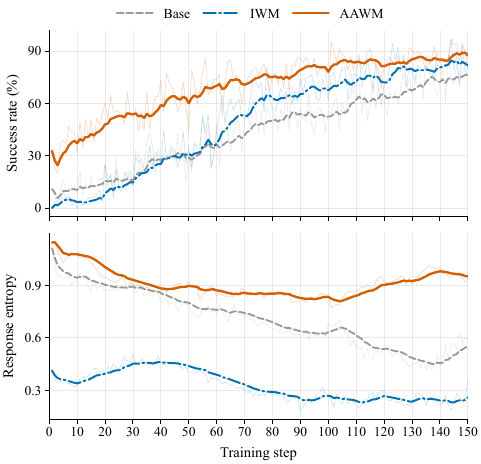}
\caption{GRPO training dynamics on ALFWorld under the imitation learning then reinforcement learning setting initialized by world modeling.
The backbone is Qwen2.5-1.5B-Instruct.
The top panel shows success rate, and the bottom panel shows response entropy.
Bold curves show time-weighted EMA smoothing with coefficient 0.95.
\textsc{AAWM} reaches higher success earlier while maintaining higher response entropy.}
\label{fig:alfworld-dynamics}
\end{figure}

\Cref{fig:alfworld-dynamics} shows the ALFWorld training dynamics under the same imitation learning then reinforcement learning setting as \Cref{fig:webshop-dynamics}.
\textsc{AAWM} reaches higher success earlier than Base and \textsc{IWM}, while also maintaining higher response entropy through most of training.
This matches the WebShop pattern and shows that the faster improvement of \textsc{AAWM} does not come from early output collapse.
Together with the behavioral measures in \Cref{tab:behavioral-exploration}, this pattern supports more success-oriented exploration during early GRPO training.

\section{Prompt Templates}
\label{app:prompts}

This appendix reproduces the prompt templates used in Self-Probing and Dynamics Synthesis
(\cref{sec:method:selfprobing,sec:method:synthesis}) and evaluation
(\cref{sec:analysis}).
Curly-brace tokens (\texttt{\{...\}}) denote runtime variables.


\begin{promptbox}{Self-Probing prompt}
\texttt{\{decision\_context\}}

\medskip
You are an agent operating in an interactive environment. Before taking
your next action, externalize your understanding --- your internal model
of how this environment operates.

\medskip
Output exactly in this format --- two categories, with the number of
statements specified below:

\medskip
{[CONFIRMED PATTERNS]}\\
Cause-and-effect relationships, requirements, or constraints you
believe to be true about how this environment operates. State them as
general rules you've inferred, not observations you currently see.

\medskip
{[OPEN QUESTIONS]}\\
Specific uncertainties about environment mechanics whose answers would
affect which action you choose next.

\medskip
Format: Each statement on its own line. No numbering, no bullet points.

\medskip
Your statements will be verified and answered, then provided back to you
to inform your current decision.

\medskip
For this environment, output exactly
\texttt{\{max\_beliefs\}} confirmed pattern(s) and exactly
\texttt{\{max\_questions\}} open question(s). Do not output extra
statements.
\end{promptbox}

\smallskip
\noindent
The variables \texttt{max\_beliefs} and \texttt{max\_questions} are set
to~$n$, the per-state proposition count reported in
\cref{sec:exp:setup}.


\begin{promptbox}{Dynamics Synthesis prompt}
\texttt{\{decision\_context\}}

\medskip
--- Agent's Current Understanding ---\\
{[CONFIRMED PATTERNS]}\\
\texttt{\{confirmed\_patterns\}}

\medskip
{[OPEN QUESTIONS]}\\
\texttt{\{open\_questions\}}

\medskip
--- Retrieved Evidence ---\\
\texttt{\{retrieved\_transition\_evidence\}}

\medskip
--- Current Step Outcome ---\\
\texttt{\{immediate\_transitions\}}

\medskip
Internalize the retrieved evidence and the current step outcomes shown
above, then perform the following task. Write as if stating knowledge
directly --- do not reference or cite the evidence records.

\medskip
Your task:\\
1. Correct any confirmed patterns that appear to be wrong.\\
2. Answer the open questions where possible.\\
3. Synthesize the key environment dynamics most relevant to the current
task goal and action selection.

\medskip
Requirements:\\
-- Describe cause-effect relationships specific to this environment ---
not general knowledge any agent would already possess.\\
-- Focus on dynamics bearing on the current task goal and the situation
at hand.\\
-- DO NOT contradict the facts observed in the current and prior
steps.\\
-- DO NOT include action suggestions or plans.\\
-- Write a concise, coherent paragraph with no preamble or closing
remarks.
\end{promptbox}


\smallskip
\noindent
The \texttt{retrieved\_transition\_evidence} field groups $K$~retrieved
transitions under the proposition that retrieved them:

\begin{promptbox}{Retrieved transition formatting}
For confirmed pattern 1:\\
\quad Record 1:\\
\qquad State: ``\texttt{\{observation\_t\}}''\\
\qquad Action: ``\texttt{\{action\_t\}}''\\
\qquad Result: ``\texttt{\{observation\_t1\}}''\\
\quad \ldots\\
\quad Record $K$:\\
\qquad State: ``\texttt{\{observation\_t\}}''\\
\qquad Action: ``\texttt{\{action\_t\}}''\\
\qquad Result: ``\texttt{\{observation\_t1\}}''

\medskip
\ldots

\medskip
For open question $n$:\\
\quad Record 1:\\
\qquad State: ``\texttt{\{observation\_t\}}''\\
\qquad Action: ``\texttt{\{action\_t\}}''\\
\qquad Result: ``\texttt{\{observation\_t1\}}''\\
\quad \ldots\\
\quad Record $K$:\\
\qquad State: ``\texttt{\{observation\_t\}}''\\
\qquad Action: ``\texttt{\{action\_t\}}''\\
\qquad Result: ``\texttt{\{observation\_t1\}}''
\end{promptbox}


\begin{promptbox}{Probing effectiveness judge prompt}
You are judging one proposition about an interactive text environment
and one synthesis paragraph.

\medskip
Input fields:\\
-- state\_context: task and current environment context.\\
-- type: P or Q.\\
-- proposition: a single confirmed pattern (P) or open question (Q).\\
-- synthesis\_output: a paragraph written after seeing transition
evidence.

\medskip
Definitions:\\
P means the agent states a dynamics pattern it believes to hold.\\
Q means the agent asks about a dynamics point it wants resolved.

\medskip
\texttt{<sample>}\\
state\_context:\\
\texttt{\{state\_context\}}

\medskip
type:\\
\texttt{\{type\}}

\medskip
proposition:\\
\texttt{\{proposition\}}

\medskip
synthesis\_output:\\
\texttt{\{synthesis\_output\}}\\
\texttt{</sample>}

\medskip
Dimensions for judgment:

\medskip
D1 Decision-Helpfulness: Output 1 if the proposition concerns an
environment mechanism, precondition, dependency, constraint, or action
effect that could meaningfully inform task completion or current/future
action choices. Output 0 if it is only a surface description, tautology,
generic observation, formatting detail, or irrelevant fact.

\medskip
D2 Synthesis Contribution:\\
-- If type=P, output 1 only when the proposition contains an error,
inaccuracy, or over-confident unsupported claim, and synthesis\_output
explicitly corrects it. Otherwise output 0, including when a correct P
is merely restated or extended.\\
-- If type=Q, output 1 only when synthesis\_output gives a definite
answer to the open question. Output 0 if it is silent, hedged, partial,
or only restates the question.

\medskip
Instruction:\\
Return strict JSON only, with fields in this order. The two score fields
must be integers, either 0 or 1:\\
\texttt{\{"rationale":"<=50 words","d1\_decision\_helpfulness":0,}\\
\texttt{"d2\_synthesis\_contribution":1\}}
\end{promptbox}

\section{Transition Retrieval Details}
\label{app:retrieval}

Each proposition $q \in P_t \cup Q_t$ from Self-Probing acts as an
independent retrieval query against the transition pool~$\mathcal{T}$
(\cref{sec:method:targets}).
Transitions are embedded with Qwen3-Embedding-0.6B and indexed with
FAISS inner-product search over normalized embeddings.
Each indexed unit is a single transition record
$(o_t, a_t, o_{t+1})$.

\noindent\textbf{Per-query MMR.}
For each proposition~$q$, we first retrieve a recall pool of~$R$
candidates by embedding similarity, then apply maximal marginal
relevance to select the final~$K$ transitions that balance relevance
to~$q$ with diversity among the selected set.
\Cref{alg:retrieval} describes the procedure.
Each proposition is processed independently; there is no
cross-query deduplication, so the same transition may appear under
different propositions.
When retrieving for a state from trajectory~$\tau_i$, transitions
originating from the same episode and overlapping step are excluded.

\begin{algorithm}[t]
\small
\caption{Per-query MMR retrieval (run independently for each
  proposition).}
\label{alg:retrieval}
\begin{algorithmic}[1]
\Require Proposition embedding~$q$; transition index~$\mathcal{T}$;
  recall pool size~$R$; final hits~$K$; trade-off~$\lambda$
\State $C \gets \text{top-}R \text{ from } \mathcal{T}$ by $\langle q, c \rangle$, excluding same-episode hits
\State $S \gets \emptyset$
\While{$|S| < K$}
  \For{each $c \in C \setminus S$}
    \State $d(c,S) \gets 0$ if $S=\emptyset$; otherwise $d(c,S) \gets \max_{s \in S} \langle c,s \rangle$
    \State $\mathrm{score}(c) \gets \lambda \langle q, c \rangle
      - (1{-}\lambda)d(c,S)$
  \EndFor
  \State $c^* \gets \arg\max_c \mathrm{score}(c)$
  \State $S \gets S \cup \{c^*\}$
\EndWhile
\State \Return $S$
\end{algorithmic}
\end{algorithm}

When $S$ is empty, the redundancy term is zero, so the first record is selected by relevance alone.

\noindent\textbf{Hyperparameters.}
\Cref{tab:retrieval-hp} lists the retrieval configuration.
The recall pool~$R$ is larger than the final~$K$ to give MMR room to
diversify within each proposition's candidate set.

\begin{table}[t]\small
\centering
\resizebox{.99\columnwidth}{!}{%
\begingroup
\setlength{\tabcolsep}{18pt}
\begin{tabular}{@{}l r@{}}
\toprule
Parameter & Value \\
\midrule
Recall pool per query ($R$) & 15 \\
Final hits per query ($K$)  & 3 \\
MMR trade-off ($\lambda$)   & 0.5 \\
Embedding model             & Qwen3-Embedding-0.6B \\
Index type                  & FAISS inner product \\
\bottomrule
\end{tabular}
\endgroup
}
\caption{Transition retrieval hyperparameters.}
\label{tab:retrieval-hp}
\end{table}

\section{Self-Probing and Dynamics Synthesis Pairs}
\label{app:probing-pairs}

This appendix presents four representative triples produced during \textsc{AAWM}
data construction.
Each case displays the task, the current observation, one confirmed pattern and
one open question elicited by Self-Probing, and the corresponding responses in
the synthesis output.
Where synthesis sentences unrelated to the displayed propositions are omitted,
the omission is marked with an ellipsis.

\begin{casebox}{ALFWorld Case 1}

\textbf{Task.}\;\;
Clean some potato and put it in microwave.

\smallskip
\textbf{Observation.}\;\;
You arrive at sinkbasin~1.
On the sinkbasin~1, you see nothing.

\tcbline

\textbf{Confirmed pattern.}\;\;
``I believe cleaning a potato with the sinkbasin~1 will remove dirt from it.''

\smallskip
\textbf{Open question.}\;\;
``I don't know if the sinkbasin~1 can clean the potato without it being on the
sink.''

\tcblower\small\RaggedRight

\textbf{Synthesis.}\;\;
The environment permits cleaning an item while holding it at the sinkbasin
without placing it on the sink first, as demonstrated by successful cleaning
actions (e.g., \texttt{clean potato 1 with sinkbasin 1}) when the sinkbasin's
contents do not include the item being cleaned.
The sinkbasin's current empty state does not block the cleaning action, and
the potato remains in inventory during cleaning, resulting in a clean potato
ready for microwave use.

\end{casebox}

\begin{casebox}{ALFWorld Case 2}

\textbf{Task.}\;\;
Clean some spoon and put it in drawer.

\smallskip
\textbf{Observation.}\;\;
You open the drawer~3.
The drawer~3 is open.
In it, you see nothing.

\tcbline

\textbf{Confirmed pattern.}\;\;
``I think the \texttt{inventory} action reveals items the agent is currently
holding.''

\smallskip
\textbf{Open question.}\;\;
``I'm unsure whether the \texttt{examine} action reveals hidden items in a
drawer.''

\tcblower\small\RaggedRight

\textbf{Synthesis.}\;\;
The \texttt{examine} action does not reveal hidden items in a drawer, as it
only confirms the visible state (e.g., ``drawer~4 is open.\ In it, you see
nothing''), which is already known after opening.
\ldots{}
The \texttt{inventory} action only displays items the agent is currently
holding, not environmental objects like spoons on countertops or in drawers,
so it cannot indicate the spoon's location without prior pickup.

\end{casebox}

\begin{casebox}{WebShop Case 1}

\textbf{Task.}\;\;
Find me machine wash men's t-shirts with long sleeve with color: black, and
size: xx-large, and price lower than \$70.00.

\smallskip
\textbf{Observation.}\;\;
Search.

\tcbline

\textbf{Confirmed pattern.}\;\;
``I think specifying price constraints in the search query will filter results
to items under \$70.''

\smallskip
\textbf{Open question.}\;\;
``I'm unsure whether including size and color in the search terms ensures they
are applied as filters.''

\tcblower\small\RaggedRight

\textbf{Synthesis.}\;\;
The search engine does not apply price constraints
(e.g., \texttt{price < 70.00}) as filters, as results include items exceeding
the price limit.
Including color and size as attributes
(e.g., \texttt{color:black}, \texttt{size:xx-large}) in the query string fails
to filter results, as non-matching items still appear.
\ldots

\end{casebox}

\begin{casebox}{WebShop Case 2}

\textbf{Task.}\;\;
Find me wide leg, slim fit, straight leg, loose fit women's shorts with color:
red, and size: xx-large, and price lower than \$40.00.

\smallskip
\textbf{Observation.}\;\;
Back to Search | {<}~Prev | color | black | blue | dark gray | gray | navy |
red | silver | size | small | medium | large | x-large | xx-large | Burband
Womens High Waist Bike Shorts Tummy Control Workout Yoga Pants Running
Exercise Running Shorts with Side Pockets | Price: \$6.82 | Rating: N.A.\
| Description | Features | Reviews | Buy Now

\tcbline

\textbf{Confirmed pattern.}\;\;
``I think color and size filters apply to the search results, not individual
product pages.''

\smallskip
\textbf{Open question.}\;\;
``I'm unsure whether changing color/size filters after clicking a product
affects the current product's display.''

\tcblower\small\RaggedRight

\textbf{Synthesis.}\;\;
The product detail page displays color and size filters that directly modify
the current product's available variants (e.g., clicking ``red'' selects the
red color option for this specific product, not the search results), while
``Back to Search'' returns to the original filtered search results without
altering the applied criteria.
This confirms that color/size selections on product detail pages are
product-specific, not global search filters, and ``Back to Search'' resets to
the search context with the same filters active.

\end{casebox}

\section{RL Rollout Reasoning}
\label{app:rollout}

During reinforcement learning, \textsc{AAWM}-initialized policies
(Qwen2.5-7B-Instruct) frequently produce reasoning traces that reference
specific environment mechanisms before selecting an action.
This appendix presents four rollout excerpts illustrating this pattern, with
the complete reasoning trace for each step.

\begin{casebox}{ALFWorld Case 1}

\textbf{Task.}\;\;
Put a hot cup in cabinet.

\smallskip
\textbf{Observation.}\;\;
You arrive at microwave~1.
The microwave~1 is closed.

\tcblower\small\RaggedRight

\textbf{Reasoning.}\;\;
The goal is to put a hot cup in a cabinet, and we have the cup in our
inventory.
To achieve this, we need to heat the cup in the microwave.
The current observation shows the microwave is closed, but the microwave can
be used to heat the cup regardless of its state (closed or open) when the cup
is in inventory.
Heating the cup in the microwave will make it hot, and we can then move it to
a cabinet.
Since we have the cup and are at the microwave, the immediate next step is to
heat the cup.

\smallskip
\textbf{Action.}\;\; \texttt{heat cup 2 with microwave 1}

\end{casebox}

\begin{casebox}{ALFWorld Case 2}

\textbf{Task.}\;\;
Put a clean spatula in diningtable.

\smallskip
\textbf{Observation.}\;\;
You arrive at sinkbasin~1.
On the sinkbasin~1, you see nothing.

\tcblower\small\RaggedRight

\textbf{Reasoning.}\;\;
Cleaning the spatula~1 with the sinkbasin~1 is a valid action that directly
transforms the spatula into a clean state.
The environment validates that this action works as expected (e.g., spatulas
are successfully cleaned when executed on a sinkbasin with them in inventory),
and the current environment state shows the spatula is in inventory and the
sinkbasin is empty (no evidence of prior cleaning attempts or water requiring
preconditions), making cleaning the immediate step to achieve the goal.

\smallskip
\textbf{Action.}\;\; \texttt{clean spatula 1 with sinkbasin 1}

\end{casebox}

\begin{casebox}{WebShop Case 1}

\textbf{Task.}\;\;
Find me gluten free, protein serving pantry staples with flavor name: flax
seeds, and size: 4.25~ounce (pack of~12), and price lower than \$70.00.

\smallskip
\textbf{Observation.}\;\;
Back to Search | {<}~Prev | flavor name | country ranch | multi-seeds |
original almond | \ldots{} | flax seeds | honey cinnamon | size | 4.25~ounce
(pack of~1) | \ldots{} | 4.25~ounce (pack of~12) | 7.7~ounce (pack of~1) |
Blue Diamond Almonds Nut Thins Gluten Free Cracker Crisps, Hint of Sea Salt,
4.25~Oz Boxes (Pack of~12) | Price: \$34.44 | Rating: N.A.\ | Description |
Features | Reviews | Buy Now

\tcblower\small\RaggedRight

\textbf{Reasoning.}\;\;
The current product listing (Blue Diamond Almonds) already satisfies the size
and price criteria (4.25~ounce pack of~12 for \$34.44), but the flavor is not
``flax seeds'' as specified in the goal.
The action to correct this is to change the flavor filter to ``flax seeds''
by clicking on the ``flax seeds'' option in the flavor menu.
This will update the product listing to show only items with the correct
flavor, ensuring the goal's flavor requirement is met.

\smallskip
\textbf{Action.}\;\; \texttt{click[flax seeds]}

\end{casebox}

\begin{casebox}{WebShop Case 2}

\textbf{Task.}\;\;
Find me loose fit women's tops, tees \& blouses with short sleeve, long sleeve
with color: c5-wine, and size: large, and price lower than \$40.00.

\smallskip
\textbf{Observation.}\;\;
Back to Search | {<}~Prev | size | small | medium | large | x-large |
xx-large | color | a1-blue | a2-black | a2-wine | \ldots{} | c5-army green |
c5-black | c5-blue | c5-gray | c5-green | c5-purple | c5-red | \ldots{}

\tcblower\small\RaggedRight

\textbf{Reasoning.}\;\;
The product ``FABIURT Summer Tops for Women'' has already been filtered by
c5-wine color, so it remains to select the size large.
The size option ``large'' is listed in the available size options and does not
conflict with the color choice.
Selecting size large will satisfy the goal of finding products in size large
while maintaining the color c5-wine.

\smallskip
\textbf{Action.}\;\; \texttt{click[large]}

\end{casebox}

\section{Use of Scientific Artifacts}
Our experiments use publicly available research artifacts.
ALFWorld, WebShop, and the AgentGym codebase are released under
MIT licenses.
The Qwen2.5-1.5B-Instruct, Qwen2.5-7B-Instruct,
Qwen3-Embedding-0.6B, and Qwen3-30B-A3B-Instruct-2507 models
are released under Apache-2.0 licenses.
The verl-agent (GiGPO) codebase is also released under Apache-2.0.
The AgentTraj-L dataset does not specify an explicit license on
its public dataset page; we therefore use it only for research
experiments and do not redistribute the original trajectories or
any derivative training targets.
Should we release code, prompts, or trained checkpoints,
we will do so under terms compatible with the licenses of the
underlying artifacts.

\end{document}